# Classifier-Based Text Simplification for Improved Machine Translation

Shruti Tyagi
Department of Computer Science
Banasthali University
Rajasthan, India
tyagi.shruti91@gmail.com

Deepti Chopra
Department of Computer Science
Banasthali University
Rajasthan, India
deeptichopra11@yahoo.co.in

Iti Mathur
Department of Computer Science
Banasthali University
Rajasthan, India
mathur_iti@rediffmail.com

Nisheeth Joshi
Department of Computer Science
Banasthali University
Rajasthan, India
jnisheeth@banasthali.in

*Abstract*— Machine Translation is one of the research fields of Computational Linguistics. The objective of many MT Researchers is to develop an MT System that produce good quality and high accuracy output translations and which also covers maximum language pairs. As internet and Globalization is increasing day by day, we need a way that improves the quality of translation. For this reason, we have developed a Classifier based Text Simplification Model for English-Hindi Machine Translation Systems. We have used support vector machines and Naïve Bayes Classifier to develop this model. We have also evaluated the performance of these classifiers.

## I. INTRODUCTION

Text Simplification is the process that reduces the linguistic complexity of the data while retaining the original text and meaning. It is the process of enhancing natural language which improves the readability and understandability of the text. Text simplification can be used in the following areas:

*A. Aphasic and Dyslexia readers:* Aphasia readers have difficulty in understanding long and complex sentences and Dyslexia readers face difficulty in understanding complex words. Text Simplification is an approach that helps in resolving these problems.

*B. Language Learners:* People having limited vocabulary faced difficulty in learning any new language. Text Simplification is a process that solves this issue.

*C. Parsing:* long sentences are difficult to parse. Using Text Simplification, the throughput of the parser can be increased.

*D. Machine Translation:* Complex sentences can be replaced by simple sentences to improve the quality of Machine Translation.

*E. Text Summarization:* Splitting of sentence or conversion of long sentence into smaller ones helps in text summarization.

There are also different approaches through which we can apply text simplification. These are:

*A. Lexical Simplification:* In this approach, identification of complex words takes place and generation of substitutes/synonyms takes place accordingly. For example, in the example below, we have highlighted the complex words and their respective substitutes.
**Original Sentence:** Audible word was **originally named** audire in Latin.
**Simplified Sentence:** Audible word was **first called** audire in Latin.

*B. Syntactic Simplification:* Syntactic Simplification is the process of splitting of long sentences into smaller ones. Given example illustrates the process.
**Original Sentence:** Jaipur, which is the capital of Rajasthan, is popularly known as the pink city and Jaipur is also a tourist place which attracts tourists from different parts of the world and which is famous for marble statues and blue pottery.
**Simplified Sentence:** Jaipur is the capital of Rajasthan. It is popularly known as the pink city. It is also a tourist place which attracts tourists from different parts of the world. It is famous for marble statues and blue pottery.

*C. Explanation Generation:* In Explanation Generation, an explanation will be provided to the complex phrases/words. Following example shows the process.
**Original Sentence:** Birth defect
**Simplified Sentence:** Pulmonary atresia (a type of birth defect).

The rest of the papers is organized as: Section II gives a brief description of the work done in the area of text simplification. Section III describes experimental setup. Section IV describes proposed methodology. Section V discusses the evaluation results and section VI concludes the paper.

## II. LITERATURE SURVEY

Machine translation is an important research field. Lot of research has already been done in computational linguistics. Siddharthan [1] presented architecture for text simplification which consists of three stages- analysis, transformation and regression and these stages have been developed and evaluated separately. In their paper, they have mainly focused on the discourse level aspects of syntactic simplification. They have considered adjectival clauses, adverbial clauses, coordinated clauses, subordinated clauses and correlated clauses to perform syntactic simplification.

Specia et al. [2] presented an approach to identify the translation quality of the target sentence by considering confidence estimation. They have used 30 black-box features and 54 glass-box features. They uniformly distributed the WMT dataset as 50% (training), 30% (validation), and 20% (testing). They compared their results with the previously developed methods and found out that they have produced better results in terms of CE score i.e. 0.602.

Aluisio et al. [3] developed a tool SIMPLIFICA that determines the readability level of the source sentences using classification, regression and ranking. With SIMPLIFICA, they have investigated the complexity of the original text. They have used two approaches i.e. natural- used to simplify the selective portions and strong- used to simplify the whole sentence. The techniques with the presence of feature set attain good performance.

Saggion et al. [4] presented an approach SIMPLEXT of text simplification for Spanish. The objective was to improve the accessibility of the text. Mirkin et al. [5] manifested a way to enhance the source text prior to translation using SORT as a web application with MVC (Model View Controller). The rewritings was generated by estimating the confidence scores of source and simplified text. 440 pairs of sentences were examined and observed that 20.6% were original, 30.4% were rewritten and rest 49% no solution.

Paetzold and Specia [6] have analyzed both syntactic and lexical simplification by learning of tree transduction rules using Tree Transduction Toolkit (T3). They have used 133K pairs of source sentence from Simple English Wikipedia corpora. They have evaluated the text automatically by using BLEU and obtained 0.342 score and manually by using Cohen's kappa and found a range 0.32(fair) and 0.68(substantial). They have concluded that the results for lexical simplification were more inspiring.

In India, some researchers also have tried text simplification. Ameta et al. [7] developed a Gujarati stemmer which they applied with a rule based for text simplification and improving the quality of Gujarati-Hindi machine translation system [8]. Patel et al. [9] presented a reordering approach. They have used Stanford parse tree at the source side. According to their approach, the source text rearranged themselves according to the structure of target text. They have evaluated the quality of text in terms of BLEU, NIST, mWER, mPER which was 24.47, 5.88, 64.71 and 43.89 respectively. They have concluded that, adding more rules for reordering automatically improves the translation quality. Narayan and Gardent [10] showed an approach comprised of deep semantic characteristics and monolingual machine translation model for text simplification using PWKP corpus. They performed both automatic evaluation using BLEU and FKG and human evaluation and compared their results with three other approaches which were produced by zhu, woodsend and wubben and found out that their approach ranked first in terms of simplicity, fluency and adequacy.

Classifier based processing has also been applied in Indian languages. Gupta et al. [11][12] developed a naïve bayes classifier through which they tried to analyze the quality of machine translation outputs so that they can be ranked. Gupta et al. [13] developed a language model based approach for ranking MT outputs which they improved by adding stemmer assisted ranking [14] and then by adding more linguistic features [15]. Joshi [16] developed a test and training corpus for training of classifier for automatic MT evaluation. Joshi et. al. [17] used this corpus in training two classifiers. A decision tree based classifier and a support vector machine based classifier and showed that using classifier based evaluation provides better correlation with human evaluation then automatic evaluation.

## III. EXPERIMENTAL SETUP

In order to train our classifiers we needed some training corpus. Thus we trained English to Simplified English machine translation system using Moses machine translation toolkit [18]. We used PWKP parallel corpus developed using Wikipedia [19]. We trained a phrase based model using this corpus. Once this was done, we collected 3000 more complex sentences and generated the outputs using the phrased based model trained on English to Simplified English. Next we asked a human annotator to manually verify if the translated simplified English output had the same meaning as the of the original sentences or not. We applied a simple binary classification where if the meaning was preserved and the output produced give simplified English sentence then Yes was given as classification and No otherwise. This comprised our training set which had complex English sentence, Its simplified translation and a binary classification (Yes/No). Table I shows the statistics of our training corpus and table II shows a snapshot of our training corpus.

TABLE I: STATISTICS OF TRAINING CORPUS

| Corpus | English-Simple English Parallel Corpus | |
|---|---|---|
| Sentences | 3000 | |
| | English | Simple English |
| Words | 68635 | 45032 |
| Unique words | 19008 | 12913 |

TABLE II: BINARY CLASSIFICATION

| English Sentence | Simple English Sentence | Binary Classification |
|---|---|---|
| We can take many means of transportation such as cars, bus or rickshaws to migrate from one place to another in Delhi. | We can take cars, bus or rickshaws to move from one place to another in Delhi. | Yes |
| January is the first month of the year according to the Hindu Calendar, and one of the seven months with 31 days. | January is the first month of the year with 31 days. | Yes |
| The birthstone of Aquarius is Amethyst, and the birth flower is Orchid flowers. Other flowers are Solomon's seal and Golden rain. | Aquarius's flower is Orchid flowers and its birthstone is the Amethyst. | No |
| February is the second month of the year according to the Hindu Calendar with the length of 28 or 29 days. | February is the second month of the year with 28 or 29 days. | Yes |

## IV. PROPOSED METHODOLOGY

Next identified 17 features for classification. These features were use by Specia et al. [2] for identification of translation quality. The 17 features used were:

1. No. of tokens in source sentence
2. No. of tokens in target sentence
3. Average source token length
4. Language model probability of trigrams in source sentence
5. Language model probability of trigrams in target sentence
6. Average target tokens present in target corpus
7. Average no. of translation according to source lexicons based on word based model with 20% or more probability.
8. Average no. of translation according to source lexicons based on word based model with 10% or more probability.
9. Percentage of low frequency source words in the training corpus.
10. Percentage of high frequency source words in training corpus.
11. Percentage of low frequency source bigrams in the training corpus.
12. Percentage of high frequency source bigrams in training corpus.
13. Percentage of low frequency source trigrams in the training corpus.
14. Percentage of high frequency source trigrams in training corpus.
15. Percentage of source words present in the corpus
16. No. of punctuation marks present in the source sentence.
17. No. of punctuation marks present in the target sentence.

Our feature extraction algorithm extracted these features. We trained two classifiers based on these features. We trained a naïve bayes classifier and a support vector machine based classifier for our study. We used naïve bayes classifier because it is simple, easy to implement classifier which is based on Bayesian theorem with strong assumptions. This classifier is used where resources are limited and it produces efficient outputs. It executes very quickly and is suitable for small text classification. We used support vector machine based classifier because it is a linear classifier that divides data into two classes using a decision boundary or a hyper plane. It produces more accurate outputs for high dimensional data. Figure 1 describes our approach.

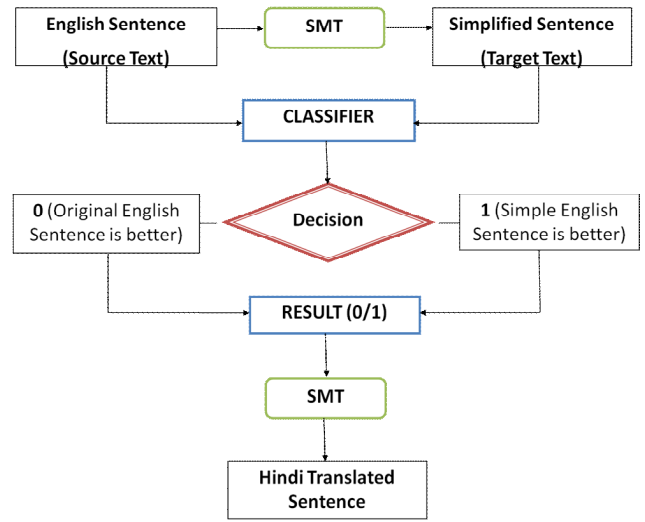

Fig. 1: Our Approach

Here, we first take the English sentence and give it to the MT engine which gives us a simple English translation of the same. These two (Original English sentence and its Simplified version) are given to the classifier for decision whether the output is a good simplification or a bad simplification. If it is a good simplified sentence then the output is sent to English-Hindi translation engine for translation otherwise original English sentence is sent for translation.

## V. EVALUATION

In order to check the performance of our classifiers, we created a test corpus of 3000 English complex sentences and gathered their simplified outputs. Next we sent these two (input complex English sentences and their Simplified English outputs) to the classifiers and registered their outputs which classified them as good or bad simplified translations. Next, we asked a human expert to do the same on two set of inputs. Based on the results obtained from the human expert and the results of the two classifiers we computed their precision, recall and f-measure scores. We also computed mean absolute error, root mean square error and kappa statistics of these two results. Table III summarized them.

TABLE III: COMPARISION OF RESULTS BETWEEN HUMAN AND CLASSIFERS

|  | Human - Naïve Bayes Classifier | Human - Support Vector Machine Classifier |
|---|---|---|
| Mean Absolute Error | 0.4618 | 0.4657 |
| Root Mean Square Error | 0.5171 | 0.6824 |
| Kappa Statistics | 0.5177 | 0.4451 |
| Precision | 0.562 | 0.527 |
| Recall | 0.565 | 0.534 |
| F-Measure | 0.563 | 0.525 |

Among these two classifiers naïve bayes classifier produced better results as compared to support vector machine based classifier. it's mean absolute error and root mean square error was lesser as compared to support vector machine based classifier. Moreover, the kappa statistics and f-measure were higher for naïve bayes classifier. F-measure is the combination of precision and recall. Even precision and recall were also higher for naïve bayes classifier. F-measure showed that 56% percent times the results of human and the classifier matched while for support vector machine based classifier, only 52% times the two results matched. To further strengthen our claims we also computed the confusion matrix of the two classifiers with the human results. Confusion matrix for naïve bayes classifier is shown in table IV and for support vector machine is shown in table V.

TABLE IV: CONFUSION MATRIX FOR NAÏVE BAYES CLASSIFIER

| Human  Machine | Yes | No | Total |
|---|---|---|---|
| Yes | 668 | 603 | 1271 |
| No | 703 | 1026 | 1729 |
| Total | 1371 | 1629 | 3000 |

TABLE V: CONFUSION MATRIX FOR SVM CLASSIFIER

| Human  Machine | Yes | No | Total |
|---|---|---|---|
| Yes | 516 | 542 | 1058 |
| No | 855 | 1087 | 1942 |
| Total | 1371 | 1629 | 3000 |

In confusion matrix of naïve bayes classifier, 1694 times both the classifier and the human agreed on the same results. Among them 668 were we good simplifications and 1026 were bad simplifications. On 1306 occasions their results did not match. Among them 603 times human adjudged the translations as bad simplified translations while the machine considered them good and on 703 occasions human considered the translations to be good, but machine did not agreed with them. In confusion matrix of SVM based classifier, 1603 times both human and machine's results matched. Among them 516 times them both agreed that the translations were good and 1087 they agreed that the translations were bad. On 1397 occasions their results did not matched. Among them, 542 times, human concluded that the translations were bad, but the machine concluded that they were good. On 855 occasions, human concluded that the translations were good but machine adjudged them as bad translations. Thus from confusion matrix it is clear that the results of the naïve bayes classifier are more accurate as compared to the results of the support vector machine based classifier.

VI. CONCLUSION

In this paper, we have developed a Classifier Based Text Simplification model for identifying if the produced results are good or bad simplified versions of the original input sentence. For this, have trained support vector machine and naïve bayes classifiers. We tested these two classifiers on 3000 sentences and found that naïve bayes classifiers has slightly better score then support vector machine based classifier. Not only we calculated precision, recall and f-measure scores which are considered as the standard evaluation measures but we also calculated mean absolute error and root mean square error and kappa statistics. Finally to strengthen our claim we verified the results with the analysis using confusion matrix.